\documentclass[a4paper, 10pt, conference]{ieeeconf}      

\IEEEoverridecommandlockouts                              

\usepackage{graphbox,graphicx}
\usepackage{gensymb}
\usepackage{graphics} 
\usepackage{epsfig} 
\usepackage{mathptmx} 
\usepackage{times} 
\usepackage{amsmath} 
\usepackage{amssymb}  
\usepackage{bm}
\usepackage{hyperref}
\usepackage{cleveref}
\usepackage{xcolor}
\usepackage{wasysym}
\usepackage{booktabs}
\usepackage{url}

\DeclareMathOperator*{\argmin}{arg\,min}

\newcommand{\vect}[1]{\boldsymbol{#1}}
\title{\LARGE \bf
Towards Very Low-Cost Iterative Prototyping for Fully Printable Dexterous Soft Robotic Hands}

\author{Dominik Bauer$^{1}$, Cornelia Bauer$^{1}$, Arjun Lakshmipathy$^{2}$, Roberto Shu$^{1}$, Nancy S. Pollard$^{1,2}$

\thanks{$^{1}$Robotics Institute, Carnegie Mellon University, 5000 Forbes Ave, Pittsburgh, PA 15213, USA. }
\thanks{${^2}$Computer Science Department, Carnegie Mellon University, 5000 Forbes Ave, Pittsburgh, PA 15213, USA.}
\thanks{\{dominikb,cornelib,aslakshm,rshu,nsp\}@andrew.cmu.edu}%

}

\begin{document}

\maketitle
\thispagestyle{empty}
\pagestyle{empty}

\begin{abstract}
The design and fabrication of soft robot hands is still a time-consuming and difficult process. Advances in rapid prototyping have significantly accelerated the fabrication process  while introducing new complexities into the design process. In this work, we present an approach that utilizes novel low-cost fabrication techniques in conjunction with design tools to help soft hand designers systematically take advantage of multi-material 3D printing to create dexterous soft robotic hands. While very low-cost and lightweight, we show that generated designs are highly durable, surprisingly strong, and capable of dexterous grasping.
\end{abstract}

\section{Introduction}
Traditionally, dexterous anthropomorphic robot hands require a large number of parts and actuators to realize complicated joint mechanics \cite{Xu2016Design,shadowhand,biomimetic_hand}. Aside from the tedious assembly, changes to form and function of individual parts are often costly, time consuming, and difficult to achieve without a complete redesign and testing of the hand. 
Furthermore, rigid robot hands require complex control and sensing strategies to overcome their lack of compliance. 

In recent years many researchers have thus started to incorporate compliant mechanisms into their designs \cite{Ma2013Modular, homberg2019robust,Santina2018Pisa}, and developed hands made entirely from soft materials \cite{Deimel2016novel,bauer2020design,Abondance2020Dexterous}. In addition to the benefits associated with simpler control through underactuation and compliance, soft robots also promise to greatly reduce the number of parts needed in such a system by replacing intricate rigid body joint mechanics with simple compliant mechanisms \cite{Odhner2014Compliant}.
However, a common disadvantage of soft robots is that manufacturing usually requires a time-consuming multi-step fabrication process that involves mold making, casting, curing and support removal \cite{pneuflex_tutorial,king2018design}.
Additionally, soft hands are inherently difficult to model, often requiring the support of specialized finite-element-method (FEM) simulation to account for complicated soft body contact dynamics and continuous deformation behavior resulting from soft materials \cite{elsayed2014finite,schlagenhauf2018control,Tawk2019Soft}. As a result, determining hand morphology and placement of actuators remains challenging and requires technical expertise, intuition, and multiple iterations of designing, fabricating, and testing hand designs \cite{teeple2021role}.

\begin{figure}[thpb]

  \centering
    \includegraphics[width=0.98\columnwidth]{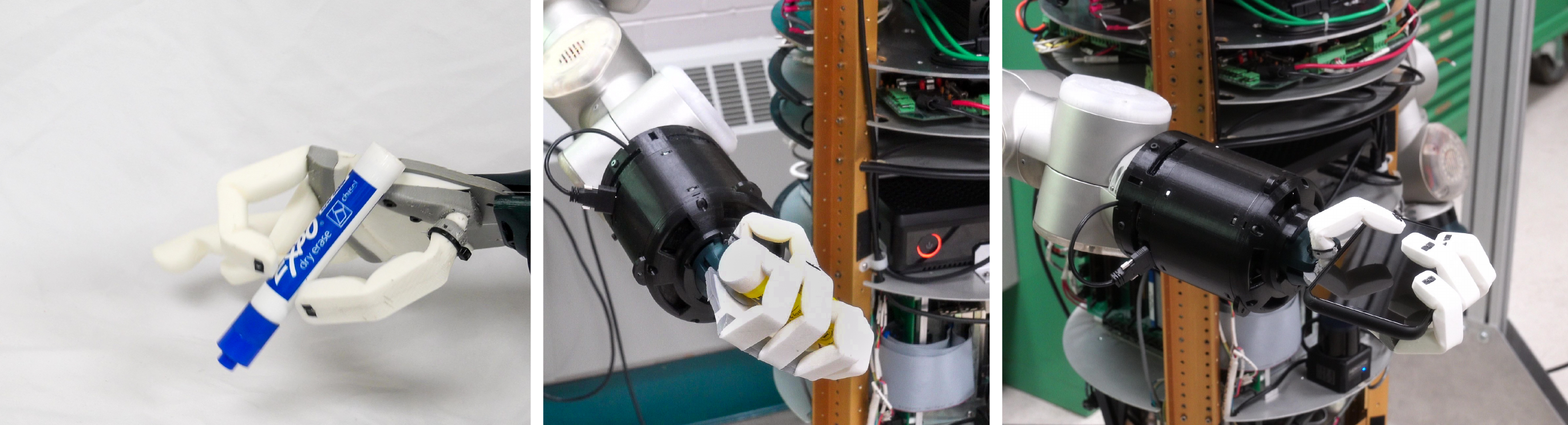}
  \caption{ Soft hand design fully printed from flexible TPU-filament performing a variety of grasps on objects of different shapes, including a precision tripod grasp holding a marker (\textit{left}), a power grasp holding a glue stick (\textit{center}), and a smartphone (\textit{right}). The hand design is self-contained including motors and can be mounted to any robot arm or humanoid robot such as the  CMU-Ballbot \cite{Shu2019Development}.  }
  \label{fig:teaser}
  \vspace{-5mm}
\end{figure}

To address these shortcomings, we make the following contributions. Firstly, we introduce a novel rapid prototyping process for tendon-actuated soft manipulators through the coupling of fast kinematic grasp simulation and design modification, effectively enabling a systematic, data-driven approach for testing the behavior of soft manipulators in relation to design changes \textit{prior} to fabrication. Secondly, we propose a low-cost fabrication process for fully printable tendon-driven soft robotic hands of variable stiffness using only low-cost fused deposition modeling (FDM) printers, commercially available flexible filaments, and off-the-shelf materials (\Cref{fig:teaser}). With upfront costs for tools and printers of less than $\$ 1000$, minimal assembly, and no post-processing requirements, we believe this process will be highly accessible to the robotics community and even hobby enthusiasts.

\section{Related Work}
\subsection{Design and Modeling}
A common disadvantage of soft robots is that modeling and simulation of designs are significantly more challenging than for traditional rigid robots. Unlike rigid hands which feature finite degrees of freedom (DoF) that can be characterized along well-defined kinematic chains, soft manipulators feature infinite degrees of freedom and a complex design space \cite{rus2015design,li2019bio}. This makes it inherently difficult for designers to predict how even small changes to design parameters such as tendon placement impacts the overall capabilities of the hand design \cite{bauer2020design,chen2020design}. Consequently, design candidates must often be fabricated and tested in the real world or require specialized simulation for evaluation \cite{elsayed2014finite,feng2018soft,Macklin2019Non-smooth,Fang2020Kinematic}. Unfortunately,  state-of-the art soft body simulators \cite{talbotsofa,ciccarelli2019particle} are not able to provide effective, efficient, and robust evaluation of design candidates for design optimization techniques \cite{chen2020design}. Further, they rely on the use of triangulated mesh geometries, making it difficult to test incremental design changes in a rapid prototyping context. Our work instead constrains deformation behavior in hand designs through the introduction of geometric features such as bumps, creases, or the combination of different materials resulting in segmented `joint-like' deformations. This enables the use of quasi-rigid approximation of traditional joints \cite{jin2020large} and thus the usage of fast optimization techniques originally intended for rigid body actuators \cite{lakshmipathy2021contact} without sacrificing the benefits of soft robots associated with compliance.

\subsection{Fabrication and Additive Manufacturing}
Widespread adoption of rapid prototyping techniques has significantly accelerated and simplified the process of designing and fabricating soft robots in recent years. 3D printing has been used to directly print molds for casting soft materials such as silicone rubber or polyurethane foam \cite{pneuflex_tutorial,king2018design,Abondance2020Dexterous}, and the development of printable soft materials has enabled researchers to directly print soft actuators and embedded sensors in one single manufacturing step \cite{Bruyas2014Combining,MacCurdy2016Printable,Shih2019Design,zhang2021bioinspired}. More recently, advances in multi-material additive manufacturing technology have further promoted the development of a new class of fully printable soft robots \cite{zhou20213d,MacCurdy2016Printable,wehner2016integrated,hussain2020Design}. For example, Hubbard et. al \cite{hubbard2021fully}  directly embed complex fluidic circuitry for pneumatic actuation of a soft robotic hand. Despite these recent advances that greatly reduce or eliminate the need for robot assembly, multi-step post-processing of the printed parts is often still required \cite{wehner2016integrated,hubbard2021fully}. In addition, current approaches are prohibitively costly, relying on ink-jet deposition (PolyJet) printing technology costing tens of thousands of dollars \cite{3dprinterpricelist}.

Our work is most similar to that of Zhou et. al \cite{zhou20213d}, who used low-cost FDM printers to print a tendon-driven upper-limb soft prosthetic that is lightweight and requires minimal assembly. However, given the application, their design process was driven primarily by matching human anatomy and restoring upper-limb functionality as opposed to more general task-based customization. Our work thus generalizes the fabrication approach of Zhou et. al \cite{zhou20213d} through the augmentation of an iterative prototyping framework and multi-material manufacturing, providing a more robust framework for arbitrary soft manipulator design objectives. 

 \section{Customizable Soft Hand Design Process}
\subsection{Iterative Design Framework}

\begin{figure}[thpb]
\vspace{2mm}
  \centering
    \includegraphics[width=0.98\columnwidth]{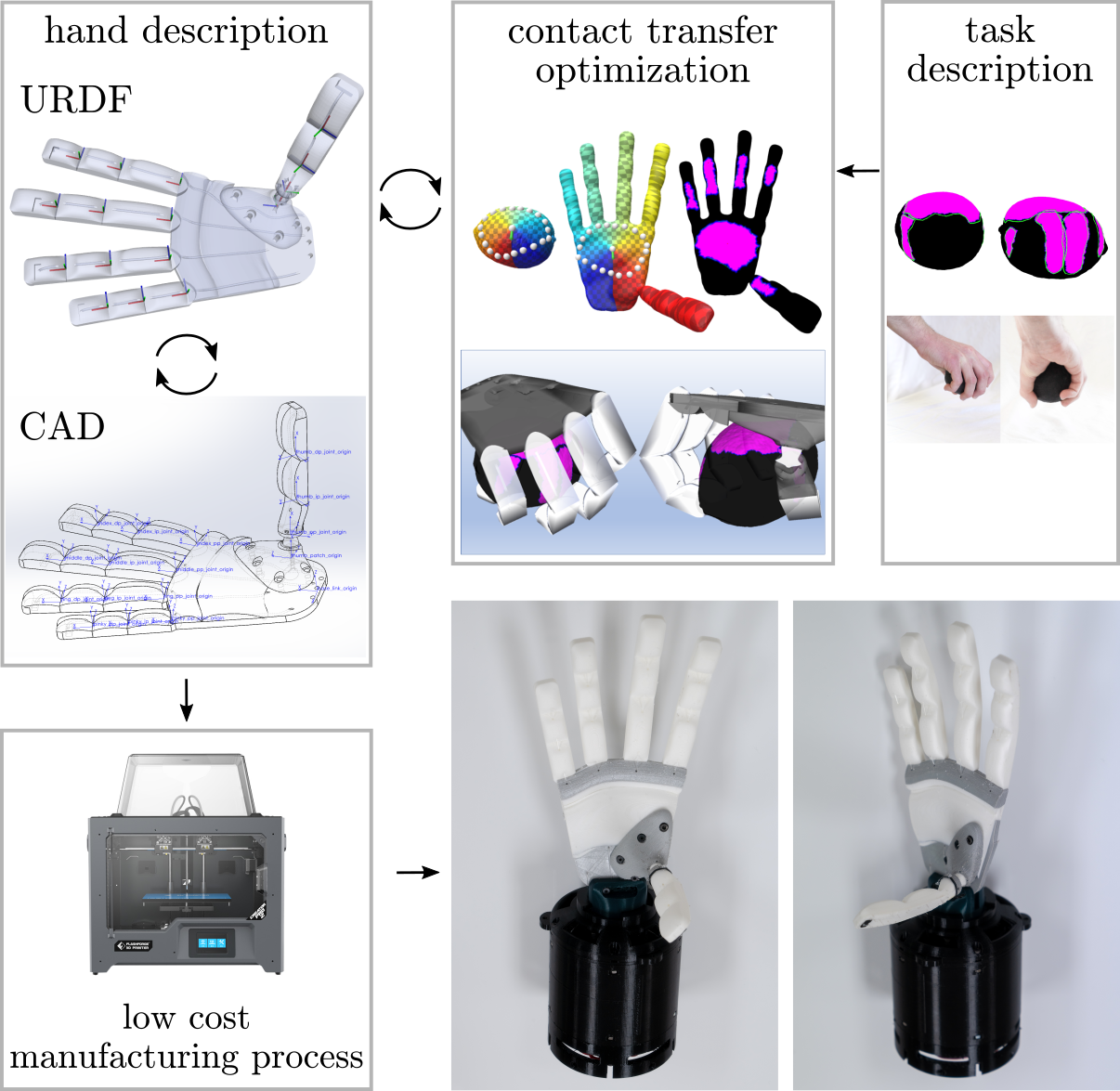}
  \caption{Our framework integrates simulation testing into the design and fabrication process of soft robotic hands. Starting from a hand design in URDF-format, we can evaluate design ideas using our contact transfer optimization approach \cite{lakshmipathy2021contact} and test design candidates against their ability to perform certain tasks recorded from human demonstrations \cite{lakshmipathy2021Tracing}. }
  \label{fig:design_framework}
  \vspace{-2mm}
\end{figure}
To aid and accelerate the design process of soft robot hands, we present a framework to generate, iterate, and evaluate hand designs for tasks recorded from human demonstration. This framework (\Cref{fig:design_framework}) builds on our previous work, which introduced a method to directly transfer grasps and manipulations performed by human subjects between objects and hands by utilizing contact areas \cite{lakshmipathy2021contact}.

We encode the high-level information of a soft hand design using the Unified Robot Description Format (URDF), which represents a joint-based description and is widely used for rigid robot hands and grippers. While this format is generally not capable of describing continuous deformations of soft materials, we show that following the design principles outlined in \Cref{sec:DesignPrinciples} allows us to model the approximate kinematic behavior of a soft hand and to quickly evaluate hand capabilities in simulation.

Our processing pipeline is depicted in \Cref{fig:design_framework}. Starting with an initial URDF model we import joint origins and axes into our computer-aided-design (CAD) software (SolidWorks, 2021). We utilize geometric features and material combinations (as detailed in \Cref{sec:DesignPrinciples}) to create flexible joints that we place along the kinematic chain to create a full CAD model of the hand.  
Then, we export mesh geometries of individual links and incorporate them into the URDF model. This can be easily done in one step using the SolidWorks URDF add-on\footnote{\href{http://wiki.ros.org/sw\_urdf\_exporter}{http://wiki.ros.org/sw\_urdf\_exporter}}. 

\begin{figure}[thpb]
\vspace{2mm}
  \centering
    \includegraphics[width=0.98\columnwidth]{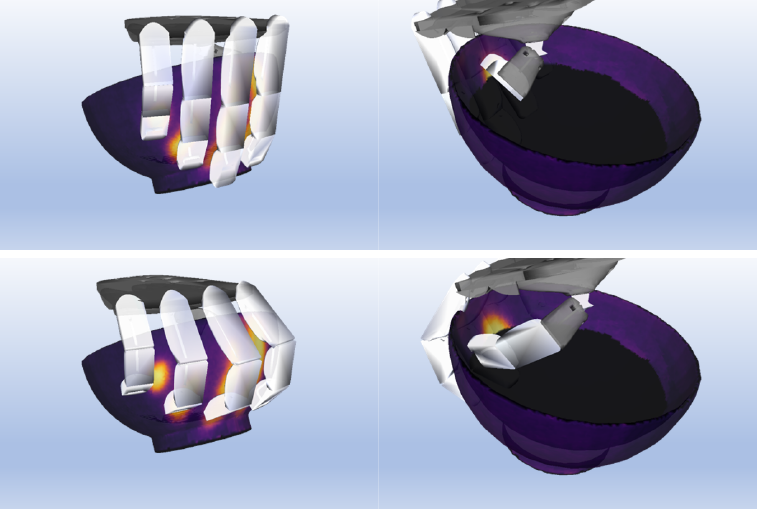}
  \caption{\textit{Top}: Solution found by our contact transfer optimization process when all rigidly approximated joint angles are independent. \textit{Bottom}: Accounting for underactuated joints that mimic independent joints produces a more realistic behavior.
  }
  \label{fig:constrained_comparison}
\end{figure}

To evaluate the generated hand design with respect to a certain task, we use our optimization-based contact transfer process \cite{lakshmipathy2021contact} to quickly synthesize kinematically feasible whole hand grasps via the following objective formulation:

\begin{equation}
    \begin{array}{rrclcl}
        \displaystyle \vect{\theta}^* = \argmin_{\vect{\theta}} & \multicolumn{3}{l}{\sum_{i=0}^{N}\ \  \Gamma_{D,i} + \lambda_n \Gamma_{N,i} + \lambda_p \Gamma_{P,i}}\\
        \mathrm{s.t.} &  \vect{\theta}_L \leq \vect{\theta} \leq \vect{\theta}_U\\
    \end{array}
    \label{eq:ctopt}
\end{equation}

where $\vect{\theta}$ is the degree of freedom vector, $\vect{\theta}_L$ and $\vect{\theta}_U$ are the lower and upper bounds of each degree of freedom respectively, $\Gamma_{D,i}$, $\Gamma_{N,i}$, and $\Gamma_{P,i}$ are the distance, normal, and prior pose deviation penalty terms for each corresponding pair of contact points $i$ respectively, and $\lambda_n$ and $\lambda_p$ are weighting hyperparameters. Detailed term explanations are available in \cite{lakshmipathy2021contact}. As demonstrated in \Cref{fig:constrained_comparison} \textit{Top}, the solutions synthesized by Eq. \ref{eq:ctopt} assume independence between all rigidly approximated joint angles, and as such do not accurately account for underactuation resulting from tendon routing. To resolve this discrepancy, we treat joints either as independently controlled or dependent, meaning that dependent joints always mimic their corresponding independent joints based on a fixed empirically determined linear relationship. During optimization, we ignore the contribution of dependent joints by setting their respective gradient contributions to zero. To ensure that proposed solutions are feasible when evaluated during objective computations, we override the dependent degrees of freedom with their corresponding linear transformations from the URDF. Importantly, we note that we deliberately treat dependent joints as soft constraints instead of introducing hard linear equality constraints. We found that hard constraints considerably escalated the problem complexity and invalidated the use of many fast gradient solvers, which combined often resulted in failure to find solutions at all.

Using our contact transfer optimization process we can quickly visualize feasible hand poses and grasps and use this visual feedback to make informed design decisions e.g. in terms of joint type, joint placement, or link lengths. Since the central representation of our approach is encoded in the URDF format, making changes to the design is straightforward.  

Overall, this framework allows us to quickly iterate through generating a variety of hand designs and evaluating their performance. 
\Cref{sec:thumb_design_study} demonstrates how we can use this process to study the influence of thumb placement and opposition on the hand's ability to achieve certain grasps.

\section{Dexterous Soft Robotic Hand Design for Multi-Material Printing}
\label{sec:DesignPrinciples}
Additive manufacturing offers key advantages over conventional fabrication techniques. Most importantly for the design of robot hands is that geometric or morphological complexity and the use of multiple materials does not result in increased manufacturing cost and complicated assembly. In this section we outline several key design principles for multi-material 3D printing which allow us to fabricate soft hand designs created within our framework. Additionally we provide examples for each of the mentioned design strategies and how they can be applied to low-cost FDM-printers. 

\begin{figure}[thpb]
\vspace{2mm}
  \centering
    \includegraphics[width=0.98\columnwidth]{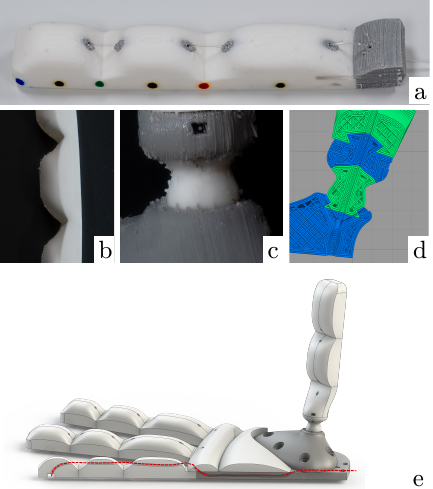}
  \caption{Design principles for creating printable, functional, and durable soft hands, including: (a) Finger with reinforced tendon-channels that improve durability. (b) Combination of bumps and creases on fingers create joint-like segmented deformation behavior. (c) Complex joints with multiple DoF using layered structures of more rigid and soft materials. (d) Interlocking boundary surfaces which prevent material separation due to excessive strain. (e) Complex arbitrary tendon channels that can be directly printed in the internal structure of the hand.}
  \label{fig:multi_material_printing_collage}
\end{figure} 
\subsection{Combining Materials with Different Properties}
Material stiffness and hand morphology typically define the deformation and kinematic capabilities in soft robot hands made from one single material.  
Combining soft materials with different properties in a single hand can be used to engineer certain deformation behaviors.
For example, printing more rigid internal bone structures can be used to form `joint-link' like kinematic chains. 
Alternatively, material properties can be locally altered to improve certain physical behaviors such as increased/decreased friction on the finger or improved abrasion resistance in moving parts. An example is provided in \Cref{sec:durability_strength}, where a more rigid material is used to reinforce the outer edges of the tendon channel to prevent the tendon from bulging out the material and eventually cutting through it after repeated actuation (see \Cref{fig:multi_material_printing_collage}-a).

\subsection{Morphology and Geometric Features}

Morphological properties such as number, length, thickness, and relative placement of fingers are central to the overall functionality of the hand design \cite{teeple2021role}. However the complex continuous design space of soft robots often makes finding the right morphology an intractable problem \cite{rus2015design}.

 
 Printed geometric features such as bumps or creases along fingers or the palm such as shown in \Cref{fig:multi_material_printing_collage}-b can be utilized to locally promote segmented planar deformation and to impose constraints on the otherwise unconstrained deformation behavior. 

For instance, a good approximation of revolute joints is given by a crease coinciding with the joint axis that runs across the finger or the palm. More complex joints with more than one DoF can be created using rigid-soft-rigid layered features where the softer material acts as cartilage to allow for deformation along multiple axes (\Cref{fig:multi_material_printing_collage}-c).

Given the typically continuous deformation behavior and complex design space of soft robots this constitutes a powerful design concept because we can approximate hand kinematics using rigid-body modeling by placing these geometric features along a kinematic chain. As a result, we are able to use existing rigid body simulators and do not need to rely on soft body dynamics or finite-element-method based simulations. 


\subsection{Actuation}
The lack of rigid structures in soft robots also poses unique challenges with respect to actuation \cite{rus2015design}. Typically actuators are either part of the overall internal structure e.g. as in the case of pneumatic actuators \cite{Deimel2013Compliant} or routed externally through anchors \cite{Calisti2011octopus} or cloth \cite{schlagenhauf2018control} in the case of most tendon(cable)-driven robots. In our approach, we show that for tendon-driven actuation it is possible to directly print internal tendon channels. This requires no post-processing or cleaning of support structures and tendons can be inserted by simply pushing them through the channel in the printed part. 
In addition to being very precise and repeatable this process allows us to incorporate complex arbitrarily shaped channels as shown in the section view of \Cref{fig:multi_material_printing_collage}-e. 
 
\subsection{Design Considerations for Multi-Material Printing with Flexible Filaments}
Printing with multiple flexible filaments and using a low-cost FDM-printer inevitably comes with limitations with respect to printability of designs and the above mentioned design principles.  
Due to the soft materials and the lack of soluble support structures, printing tall parts upright becomes difficult because the material starts to deform with the extruder once a certain ratio between part footprint and height is exceeded. To avoid this issue we design all of the hands and fingers presented in this paper such that they can be printed lying flat on the printbed. 

Considering the different material elongation properties and the large elastic deformations that materials can undergo during actuation, materials can easily separate. This is especially prevalent when soft and more rigid materials are split along a flat or smooth plane that runs orthogonal to the neutral bending plane. To improve adhesion we thus create interlocking boundary surfaces that create a form closure between materials as shown in \Cref{fig:multi_material_printing_collage}-d.




\section{Materials and Fabrication}
The soft hand presented in this work is fabricated entirely from low-cost off-the-shelf materials. All mechanical parts are printed using a desktop FDM-printer with independent dual extrusion from Flashforge (Flashforge Creator Pro 2). In order to print soft and highly flexible materials of varying  shore hardness (75A--95A)  we retrofit the printer with extruders from Flexion\footnote{\hyperref[https://flexionextruder.com/shop/dual/]{https://flexionextruder.com/shop/dual/}}. 

We use NinjaTek Chinchilla \cite{chinchilla} (Shore Hardness 75A) and Cheetah \cite{cheetah} (Shore Hardness 95A) filaments to print the hands and fingers presented in this paper, both of which are commercially available thermoplastic polyurethane (TPU) filaments. The 3D printing parameters that were used to print our multi-material prints are listed in \Cref{tab:print_params}. We export all CAD parts as STL files and slice them using the Simplify3D slicer.
This commercially available slicer is chosen over common Open-Source slicers because it allows to specify extrusion multipliers for individual extruders, a feature we find is essential when printing filaments with very different material elongation properties. All other parts are printed with standard PETG filaments using default print settings. 

For better printablity and customizability we print the thumb design separately and  mount it onto the hand using four small screws. In total the hand is actuated by seven tendons. Each finger in the hand features one flexor tendon, with the thumb featuring two additional tendons for adduction/abduction. Tendons are routed through \diameter 1 mm wide channels that are printed directly inside the hand and are made from standard monofilament fishing line with a diameter of $0.61mm$ and a rated tear strengh of $178 N$. To secure the tendon we print small dumbbell shaped anchors from PETG material and tie the tendon around the anchor using an improved clinch knot\footnote{\hyperref[https://www.animatedknots.com/improved-clinch-knot]{https://www.animatedknots.com/improved-clinch-knot}}.

Each tendon is driven by a brushless DC (BLDC) electric motor from IQ motion (IQ Vertiq 220KV\footnote{\hyperref[https://www.iq-control.com/vertiq-2306-220kv]{https://www.iq-control.com/vertiq-2306-220kv}}) which are operated at 24V. The motors feature a built-in minimum jerk trajectory generator which we use to generate smooth motions between keyframed poses. All electronic and hardware parts are contained in a compact lightweight wrist design (total weight: $648g$, soft hand excl. wrist: $94g$),  which can be mounted on robot arms (as shown in \Cref{fig:teaser} \textit{center and right}). The assembled hand design is fully self-contained requiring only a USB-cable for Serial Communication and a 24V DC power supply. 

Depending on the size of the hand a full print can take between $12-18$ hours due to the relatively slow printing speeds of flexible filaments. Once finished printing, assembling the hand is a matter of attaching the hand to the wrist, inserting the tendons through the channels and securing them by tying a simple knot. This all can be done in under one hour.

Each hand can be fabricated independently of the wrist for less than $\$5$ in costs for TPU and PETG filaments. The wrist design costs $<\$800$ in parts with servo motors ($\$94$ per motor) making up for the bulk of the cost.
The total upfront costs for the desktop 3D printer ($\$599$), custom extruders ($\$249$), and tools (soldering iron, pliers, screwdrivers etc.) are below $\$1000$, making this fabrication process less expensive than popular smartphones at the time of writing. 

On our project website\footnote{\href{https://sites.google.com/andrew.cmu.edu/rs22-printable-soft-hands}{https://sites.google.com/andrew.cmu.edu/rs22-printable-soft-hands}} we provide STL files for printing our hand design and more in-depth instructions on the fabrication process.
\begin{table}[]
    \centering
    \caption{Print parameter settings used in Simplify3D slicer for a Multi-Material TPU print with Ninjatek Cheetah and Chinchilla.}
    \begin{tabular}{l c c c}
    \toprule

\textbf{Layer} & & &\\\hline
Primary Layer Height & mm&\multicolumn{2}{c}{ 0.15} \\
Top Solid Layers & -  & \multicolumn{2}{c}{4}\\
Bottom Solid Layers & - & \multicolumn{2}{c}{4}\\\hline
\textbf{Infill } & & &\\\hline
Infill Percentage & \%& \multicolumn{2}{c}{20} \\
Infill Pattern  & - & \multicolumn{2}{c}{Rectilinear}\\\hline

\textbf{Temperature} & & &\\\hline
Extruder Temperature  & \degree C & \multicolumn{2}{c}{235}\\
Heated Bed Temperature  & \degree C & \multicolumn{2}{c}{40}\\\hline
\textbf{Speed} & & &\\\hline
Default Printing Speed& mm/s & \multicolumn{2}{c}{15.0}\\
Outline Underspeed  & \% & \multicolumn{2}{c}{80}\\
Solid Infill Underspeed  & \% & \multicolumn{2}{c}{95}\\
Support Structure Underspeed  & \% & \multicolumn{2}{c}{80}\\
X/Y Axis Movement Speed& mm/s & \multicolumn{2}{c}{50.0}\\
Z Axis Movement Speed& mm/s & \multicolumn{2}{c}{16.7}\\\hline
\textbf{Additions} & & &\\\hline
Use Ooze Shield &- & \multicolumn{2}{c}{True}\\
Offset from Part & mm & \multicolumn{2}{c}{2.0}\\
Ooze Shield Outlines &- & \multicolumn{2}{c}{1}\\\hline
 \textbf{Extrusion} & & Cheetah   &Chinchilla\\ \hline
 Extrusion Multiplier& - & 1.05&1.20 \\
 Extrusion Width & mm& 0.4 &0.4\\
 Retraction Distance &mm&1.0&0.0\\
 Retraction Speed &mm/s & 20.0 &0.0\\\hline

  \bottomrule
    \end{tabular}
    
    \label{tab:print_params}
\end{table}

\section{Experiments}
\subsection{Durability and Strength}\label{sec:durability_strength}
\begin{figure}[thpb]
\vspace{2mm}
  \centering
    \includegraphics[width=0.98\columnwidth]{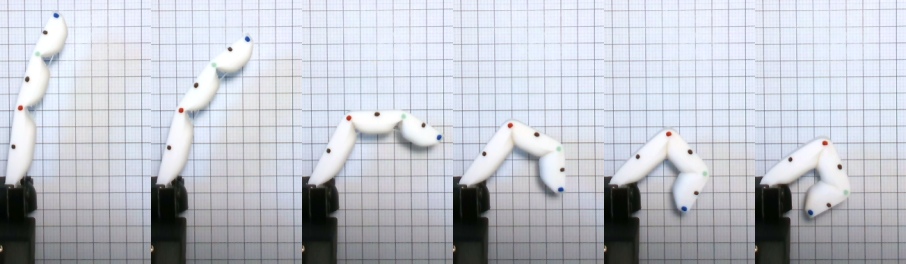}
  \caption{Durability experiment setup, from left to right: A single finger moves from a fully extended configuration to a fully flexed configuration by pulling the tendon.}
  \label{fig:durability_setup}
\end{figure}

\begin{figure}[thpb]
\vspace{2mm}
  \centering
    \includegraphics[width=0.98\columnwidth]{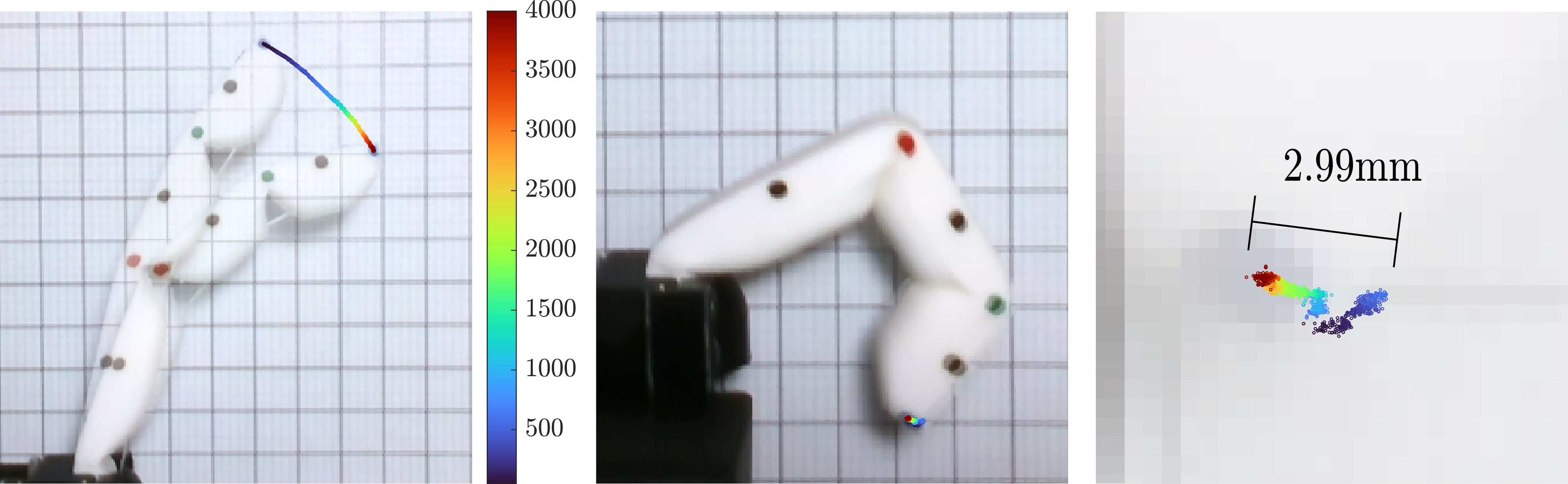}
  \caption{Durability experiment. Fingertip position in extended (\textit{left}) and fully flexed (\textit{center}) configuration over time. Overlayed images of iteration 1 and 4000 are shown, fingertip position is marked in color corresponding to iteration.
  \textit{Right}: Close-up of fingertip position in flexed configuration over time. Throughout the experiment, the fingertip position varies by a maximum of $2.99mm$.}
  \label{fig:durability_fingertip_time}
\end{figure}

\begin{figure}[thpb]
\vspace{2mm}
  \centering
    \includegraphics[width=0.98\columnwidth]{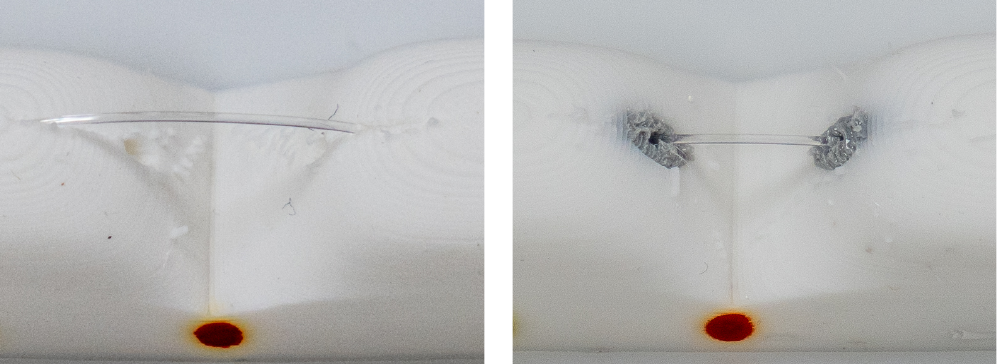}
  \caption{\textit{Left}: Damage to the printed tendon channels caused by 5200 cycles of excessive tendon actuation. \textit{Right}: Printed rigid material inserts reinforce the tendon channels. No damage is visible after 5000 cycles.}
  \label{fig:durability_tendon_channel}
\end{figure}

To investigate the durability and strength of the printed soft fingers we clamp one individual finger printed from Ninjatek Chinchilla \cite{chinchilla} and actuate the finger repeatedly until exhaustion. A Dynamixel XM430-W210-R\footnote{\href{https://www.robotis.us/dynamixel-xm430-w210-r/}{https://www.robotis.us/dynamixel-xm430-w210-r/}} servo is used to actuate the tendon. The finger weighs $10g$ and features three creases that act as `revolute-like' joints allowing for bending along the axis of the crease. We keep the tendon contraction length constant and observe material behavior by tracking colored markers on the joints using an Intel RealSense D435 camera \cite{keselman2017intel}. 
The resulting finger motion is shown in \Cref{fig:durability_setup}. The first 50 iterations are considered a \textit{break-in} period, during which any leftover debris inside the tendon channels is cleared and smoothed out by the tendon movement, and our evaluation begins after this period. We plot the fingertip position in the extended and the fully flexed configuration over experiment iterations (\Cref{fig:durability_fingertip_time}) and show overlayed images of iteration 1 and 4000.
Fingertip positions are marked in color corresponding to iteration. Hysteresis of the fully extended finger configuration is depicted in \Cref{fig:durability_fingertip_time} \textit{left}. In the fully flexed configuration (\Cref{fig:durability_fingertip_time} \textit{center, right}) fingertip positions are largely consistent over time, with a maximum distance of $2.99mm$ between positions throughout the experiment. 
No visible damage to the material is found after 4000 flexion and extension motions. 
To stress test the material's ability to withstand excessive strain, we conduct a second experiment where a finger is repeatedly actuated beyond its fully flexed configuration (\Cref{fig:durability_setup} \textit{far right}) until exhaustion. After 5200 iterations, the motion deviates significantly from the initial motion as the tendon unwraps from the pulley during extension and becomes stuck between the motor casing.
Over time we observe that the tendon slowly cuts through the fingertip patch material as shown in \Cref{fig:durability_tendon_channel} \textit{left}, causing small changes in finger motion and final flexed configuration. This increases friction between tendon and finger patch, resulting in hysteresis. 

Based on these results, we test a finger where tendon channels are coated with a thin layer of harder material  (NinjaTek Cheetah)  under the same conditions  and find that the revised design shows no signs of cutting or material fatigue after 5000 iterations as depicted in \Cref{fig:durability_tendon_channel} \textit{right}.
 
\begin{figure}[thpb]
\vspace{2mm}
  \centering
    \includegraphics[width=0.95\columnwidth]{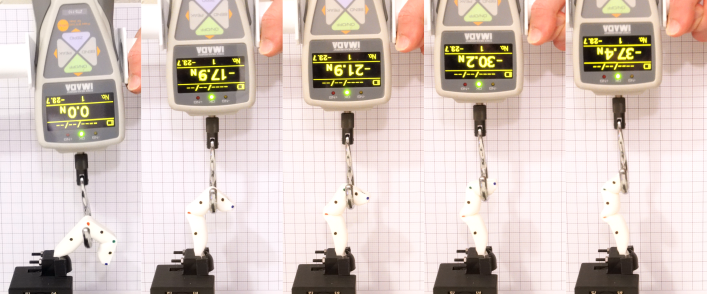}
  \caption{Individual finger strength test: In the flexed configuration, the finger withstands a maximum load of 37.4N before slipping. No visible damage to the finger was observed when examining the finger after the test.  }
  \label{fig:strength_finger}
\end{figure}

 To evaluate the strength of individual fingers, a pull-out force test is carried out as shown in \Cref{fig:strength_finger}. A hook attachment is grasped by a flexed finger, and the force required to cause pull-out is measured using a force gauge. The finger withstands a maximum load of $37.4$N before slipping, and no visible damage to the finger is observed.

 \subsection{Determining Weighting Coefficients of Dependent Joint Angles}
\begin{figure}[thpb]
  \centering
    \includegraphics[width=0.49\columnwidth]{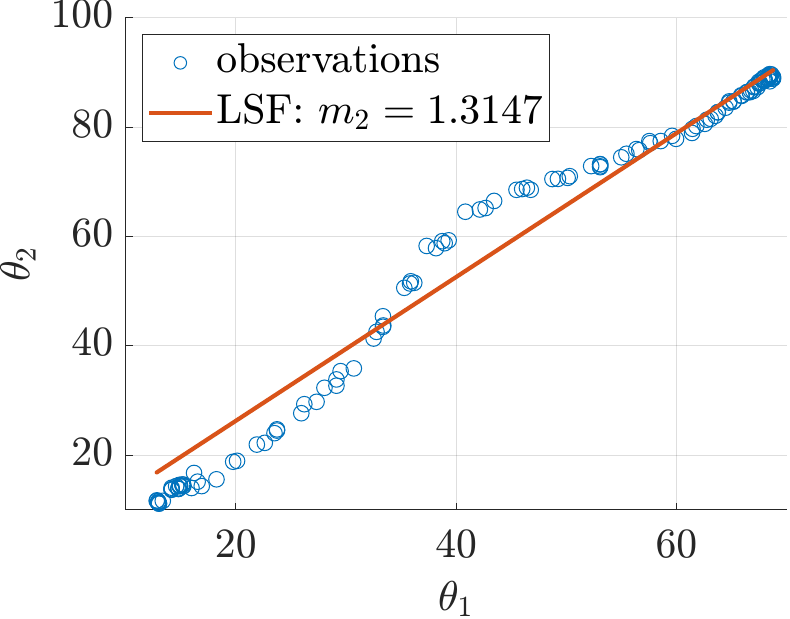}
    \includegraphics[width=0.49\columnwidth]{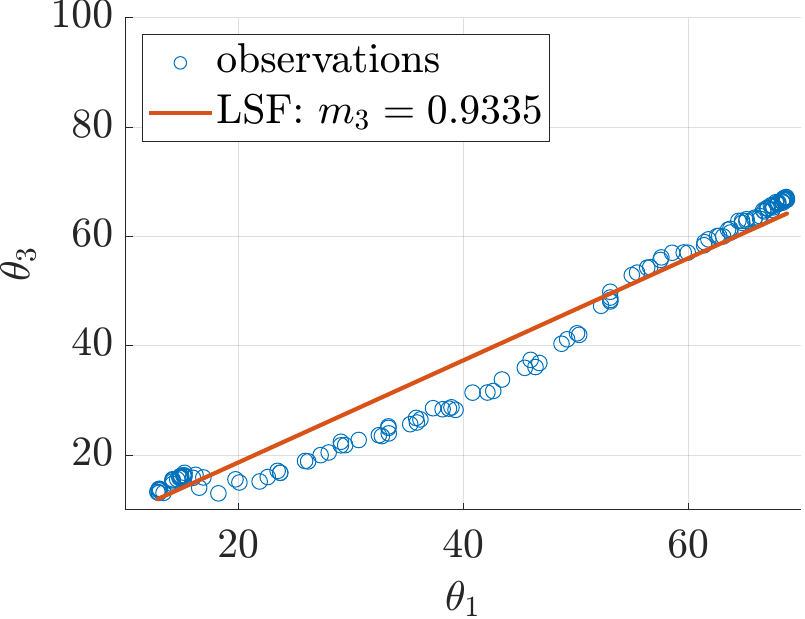}\hfill
    \vspace{-2mm}
  \caption{Joint angle trajectory data obtained from 5 recorded flexion motions of an individual finger are used to find weighting coefficients through a least squares fit.}
  \label{fig:lsf_underactuation}
\end{figure}

 When contracting tendons we observe a proportional relationship between joint angles of the same finger. We account for this underactuation in our optimization approach by introducing
 empirically determined weighting coefficients and define the proximal joint of each finger as the \textit{independent} DoF. Intermediate joints $\theta_2$ and distal joints $\theta_3$ mirror the corresponding proximal joint angle $\theta_1$, scaled by weighting coefficients. 
The relationship between joint angles $(\theta_1, \theta_2, \theta_3)$ is therefore given by $\theta_2 = m_2 \theta_1$,  $\theta_3 = m_3 \theta_1$.
We find these coefficients  through a least squares fit using joint angle trajectory data obtained from 5 recorded flexion motions of an individual finger. Data points and fitted model are shown in \Cref{fig:lsf_underactuation}.

\subsection{Design Study}\label{sec:thumb_design_study}

 \begin{figure*}[thpb]
  \centering
    \includegraphics[width=0.98\linewidth]{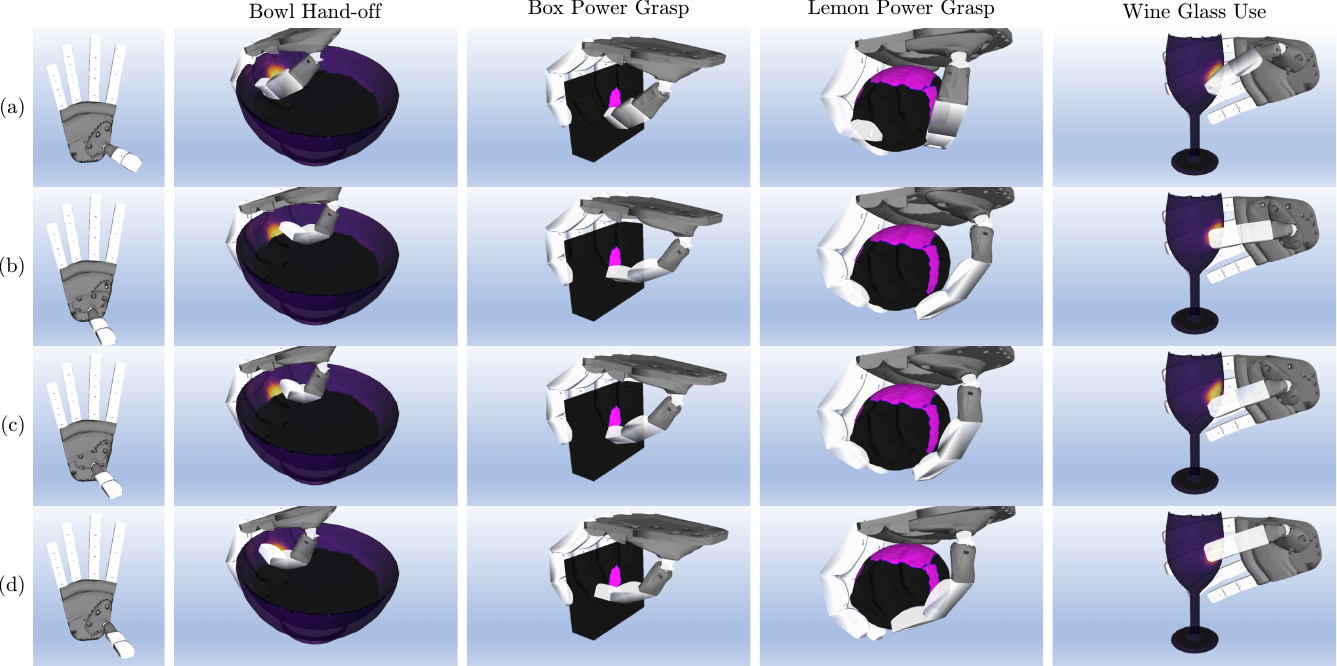}
  \caption{Anthropomorphic hand designs featuring a variety of different thumb placements (\textit{column 1}) are evaluated for their ability to grasp a bowl (\textit{column 2}), a box (\textit{column 3}), a lemon (\textit{column 4}), and a wine glass (\textit{column 5}). }
  \label{fig:thumb_design_study}
\end{figure*}

We demonstrate how our approach can be utilized for the design of anthropomorphic hands by investigating the thumb placement and its orientation on the hand. Starting from an initial predefined anthropomorphic hand design (\Cref{fig:thumb_design_study}-a), we create three different candidate designs that feature varying thumb locations (\Cref{fig:thumb_design_study}-b,c,d) and evaluate them based on grasping a bowl, a box, a lemon, and a wine glass. 

The initial design (a) produces reasonable solutions in simulation for all grasps; however, one issue becomes apparent with regards to the thumb placement. Due to the angled opposition of the thumb, most contacts are made using only the thumb's inner edge. This especially constitutes a problem for the box and lemon grasps where it becomes difficult to apply a normal force onto the object's contact region that counters the force applied by the other fingers. Further flexion of the thumb to increase contact forces would likely result in the thumb sliding upwards on the object and the hand not being able to achieve a stable grasp. We observe that this phenomena indeed renders the physical hand incapable of producing a stable power grasp on the box, which can be viewed in the supplementary video.

To create variations of our initial thumb design that are able to better apply normal forces when in contact with the contact regions of the object, we perform simple translations and/or rotations of the thumb patch directly in the URDF. In the resulting design candidates (b,c,d) the thumb visibly makes contact in a more favorable fashion, as indicated by its proximity and orientation towards the object thumb contact patch.

  \begin{figure}[thpb]
  \centering
    \includegraphics[width=0.98\linewidth]{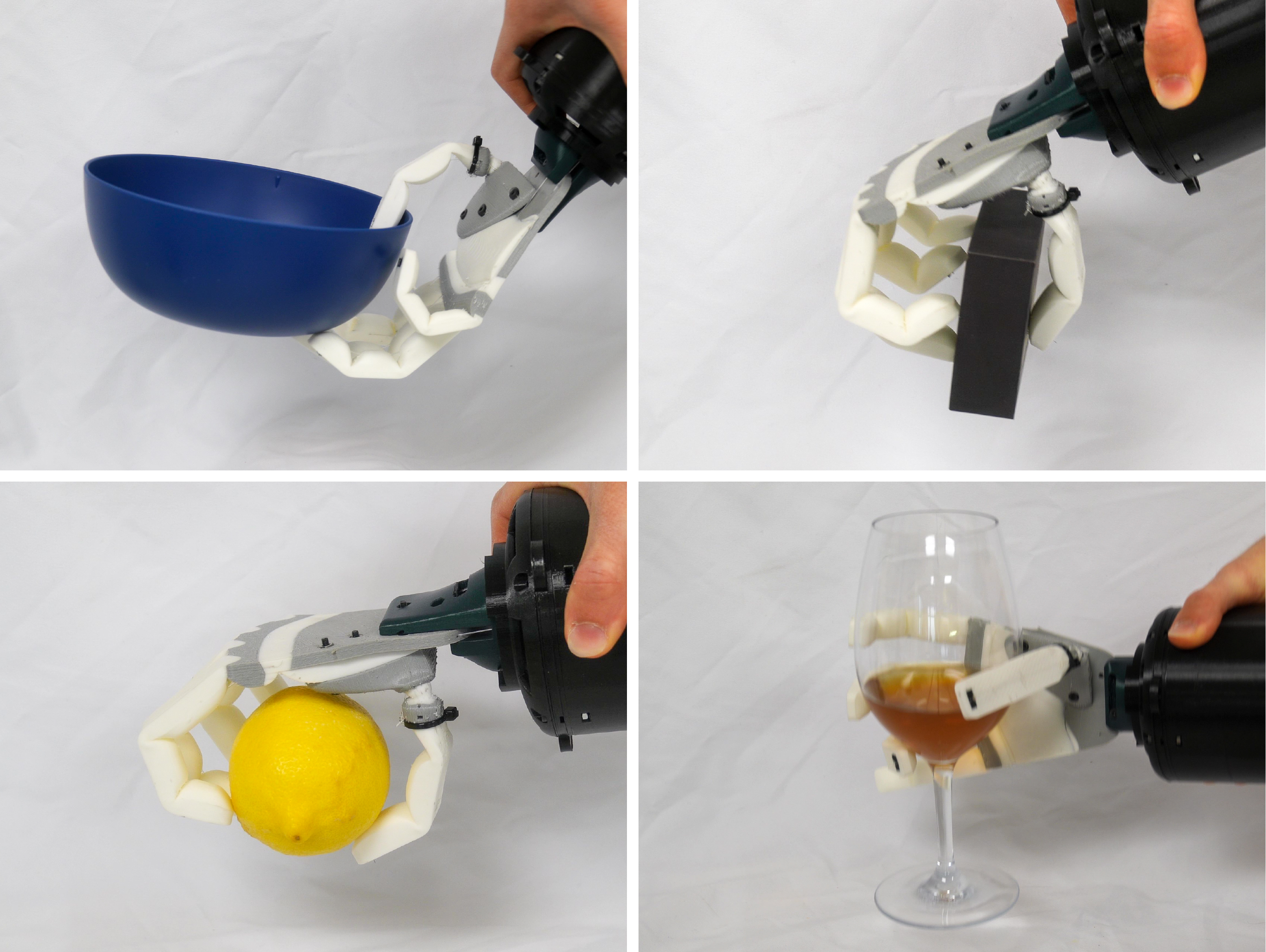}
  \caption{Based on the results in the design study shown in \Cref{fig:thumb_design_study}, candidate (d) is selected for fabrication. All grasps are successfully completed in a pick and place setup using key-framed open-loop poses that match the results obtained in our simulation.}
  \label{fig:thumb_design_revision_d}
\end{figure}
The strictly opposing thumb and finger alignments from candidates (b) and (d) generally produce more desirable contact behaviors than (c), especially for surfaces of high curvature (e.g. wineglass and lemon). However, while candidate (b) appears to better match the human demonstrated contacts, the quality of closure appears similar to candidate (d) across the simulated tasks. Due to the difference in position from the human hand - specifically the lower placement on the palm - we hypothesize that this candidate would not be able to adequately perform pinch or tripod grasps. We ultimately select candidate (d) for fabrication, and demonstrate its viable use in all four tasks depicted in \Cref{fig:thumb_design_revision_d} as well as an additional marker tripod grasp (\Cref{fig:teaser}). In particular, we note that the revised candidate was successfully able to grasp the box while the initial design failed. All real robot grasps were executed using key-framed open-loop poses that match the feasible grasps obtained in simulation. Demonstrations of the four grasps can be found in the supplementary video.

\section{Conclusion and Future Work}
Designing soft robot hands is still a time-consuming and difficult process. Advances in rapid prototyping have accelerated the fabrication process significantly, while introducing new complexities into the design process. In this work we presented an approach that utilizes novel low-cost fabrication techniques in conjunction with design tools helping soft hand designers to systematically take advantage of multi-material 3D printing. Our approach tightly integrates simulation testing with the fabrication process, allowing designers to better understand how design changes or the introduction of new design features will influence the hand's kinematic capabilities. 
We also showed that our low-cost fabrication process yields durable, robust hand designs that require little assembly and can perform a variety of dexterous grasps.  

Although our methodology provides a streamlined process to make iterative design changes most of the steps presented still require human intervention by the designer, including interpreting optimization results, slicing and preparing parts for printing, or choice of materials. In part this is the case because designing robot hands includes multiple objectives which are inherently subjective. To further automate the process, our future work will address the discovery of new design features through optimization and develop new methods to better quantify design objectives.  

Additionally the hand design presented in this work is not final but will be further improved upon with regards to kinematic capabilities, sensing, and control. To address the clutching behavior of fingers which can be observed in the design study in both the simulated and the real hand, we plan to add more than one tendon to each finger. With regards to the fabrication process, we intend to incorporate resistive or capacitive printed sensors made directly from conductive filament and use this sensory feedback to develop closed-loop control strategies.

\label{sec:Experiments}


\addtolength{\textheight}{-1.20cm}   





\section*{ACKNOWLEDGMENT}
 This material is based upon work supported by a fellowship from CMU’s Center for Machine Learning and Health awarded to Dominik Bauer and by the AI Research Institutes program supported by NSF and USDA-NIFA under AI Institute for Resilient Agriculture, Award No. 2021-67021-35329 and NSF award CMMI-1925130.
\sloppy
\bibliographystyle{IEEEtran}
\bibliography{references}

\end{document}